\documentclass[letterpaper, 10 pt, conference]{ieeeconf}  

\IEEEoverridecommandlockouts                              

\usepackage{cite}
\usepackage{graphics} 
\usepackage{epsfig} 
\usepackage{times} 
\usepackage{amsmath} 
\usepackage{amssymb}  
\usepackage{multirow}
\usepackage{color}
\usepackage{xcolor}
\usepackage{colortbl}
\usepackage{tabulary}
\usepackage[normalem]{ulem}
\usepackage{booktabs}
\usepackage{makecell}
\usepackage{algpseudocode}
\usepackage{algorithm}

\newcommand{\loss}[0]{\mathcal{L}}
\newcommand{\rot}[0]{\mathbf{R}}
\newcommand{\trans}[0]{\mathbf{t}}
\DeclareMathOperator{\avg}{avg}

\def\eg{\textit{e.g.}}

\def\ie{\textit{i.e.}}

\title{\LARGE \bf
LanPose: Language-Instructed 6D Object Pose Estimation \\ for Robotic Assembly
}

\author{Bowen Fu$^{1}$, Sek Kun Leong$^{1}$, Yan Di$^{2}$, Jiwen Tang$^{1}$ and Xiangyang Ji$^{1}$ 
\thanks{$^{1}$Department of Automation and BNRist, Tsinghua University, Beijing, China. }
\thanks{$^{2}$Technical University of Munich (TUM), Munich, Germany.}
}

\begin{document}

\maketitle
\thispagestyle{empty}
\pagestyle{empty}

\begin{abstract}

Comprehending natural language instructions is a critical skill for robots to cooperate effectively with humans. 
In this paper, we aim to learn 6D poses for robotic assembly by natural language instructions. 
For this purpose, \textit{Language-Instructed 6D Pose Regression Network} (LanPose) is proposed to jointly predict the 6D poses of the observed object and the corresponding assembly position. 
Our proposed approach is based on the fusion of geometric and linguistic features, which allows us to finely integrate multi-modality input and map it to the 6D pose in $SE(3)$ space by the cross-attention mechanism and the language-integrated 6D pose mapping module, respectively.
To validate the effectiveness of our approach, an integrated robotic system is established to precisely and robustly perceive, grasp, manipulate and assemble blocks by language commands. 
98.09 and 93.55 in ADD(-S)-0.1d are derived for the prediction of 6D object pose and 6D assembly pose, respectively. 
Both quantitative and qualitative results demonstrate the effectiveness of our proposed language-instructed 6D pose estimation methodology and its potential to enable robots to better understand and execute natural language instructions. 

\end{abstract}
\section{Introduction} \label{introduction}

Assembling blocks by language instructions is quite spontaneous for kids. 
The target blocks are first identified and manipulated, and then assembled with tight tolerances. 
In this process, the natural language instructions, \eg, \textit{insert to} and \textit{stack on}, need to be comprehended and mapped to a specific action. 
One approach to enable robots to conduct similar tasks is through the integration of geometric and linguistic information, which allows for the prediction of a precise presentation in 3D space, specifically, the 6D object pose. 
In this work, we focus on learning 6D object poses from natural language instructions to assemble blocks by a robotic system. 

Thanks to the integrative development of computer vision (CV) and natural language processing (NLP), the fusion of image and natural language information is possible. 
Specifically, visual grounding tasks \cite{kazemzadeh2014referitgame, mao2016generation, yu2016modeling} provide the 2D localization information referring to the language expression. 
Several works introduce visual grounding to robotic manipulation tasks. 
\cite{hatori2018interactively, shridhar2018interactive} both perform human-robot interaction by natural language instructions.
On the other hand, \cite{ahn2022visually} emphasizes the history-dependent manipulation to preferably conduct pick-and-place tasks. 
However, the objects are merely located by 2D center points or 2D bounding boxes. 
Without the information in 3D space with full 6 degrees of freedom, the precise assembly operation cannot be executed. 

The key challenge lies in reasoning the 6D object pose from both image and natural language information. 
The geometric features reflect the localization and orientation of the perceived target object, which contains the surrounding context information of the assembly position. 
Meanwhile, the linguistic features reveal the desired interaction between two objects, in which the relative relationship can be obtained. 
Recently, \cite{cheang2022learning} proposes to learn to grasp category-level objects by a sentence that describes the location of the objects. 
Though 6D poses are predicted for grasping, the image-text integration is only at 2D detection stage. 
Thus, only 2D location information is extracted from the text, rendering the higher-level interaction between objects impossible. 
To accomplish high-precision robotic assembly tasks, the network has to extract and integrate the image and text features, and map them to the 6D pose in $SE(3)$ space. 

\begin{figure}
	\centering
	\vspace{2mm}
	\includegraphics[width = 0.99\linewidth]{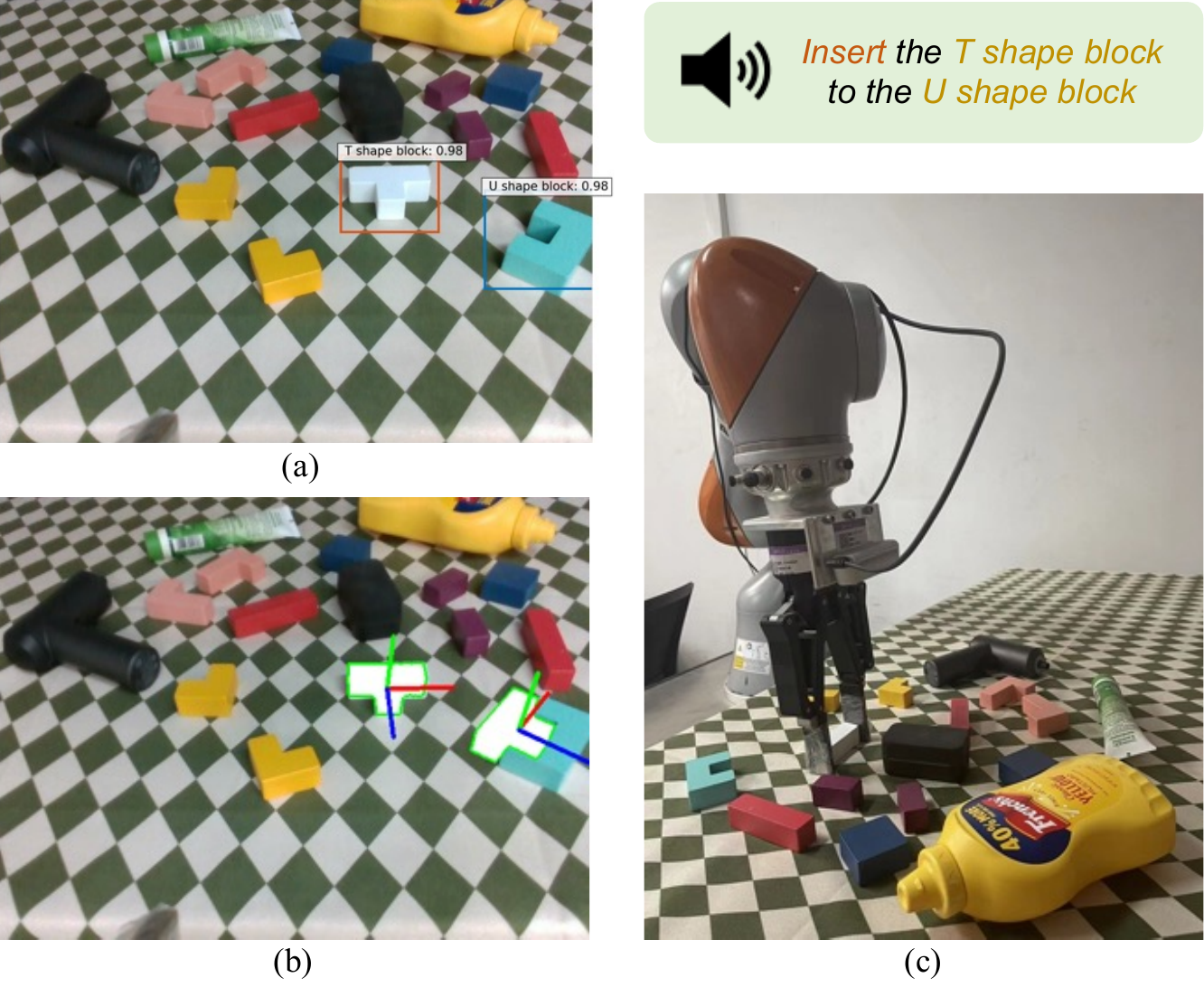}\vspace{-2mm}
	\caption{Example of our framework. Given a natural language instruction, (a) the 2D localization of target objects and (b) the 6D poses of the target objects and corresponding assembly position are predicted. (c) Our integrated robotic assembly system. }
	\label{fig:intro}
\end{figure}

In this work, we propose \textit{Language-Instructed 6D Pose Regression Network} (LanPose) to predict the 6D assembly position pose and establish an integrated robotic system to assemble blocks by a single-sentence natural language instruction (Fig.~\ref{fig:intro}). 
Fed with the target object localization from 2D visual grounding network, the LanPose integrates geometric and linguistic features by an attention mechanism to simultaneously estimate the 6D poses of the observed object and the target assembly position. 
Subsequently, an integrated robotic system is established to precisely and robustly grasped, manipulated and assembled the target object to the instructed position. 
Our networks take only RGB input and are solely trained with synthetic data, which avoids time-consuming real-world annotations, rendering our system extensible and flexible. 
We achieve 98.09 and 93.55 in ADD(-S)-0.1d in 6D pose estimation for perceived objects and assembly position and a 82.1\% success rate in real-world robotic assembly experiments. 

Our contributions can be summarized as follows: 

\begin{itemize}
\item We present \textit{Language-Instructed 6D Pose Regression Network} (LanPose) to jointly predict high-precision 6D poses of the observed object and the corresponding assembly position, which derives 98.09 and 93.55 in ADD(-S)-0.1d, respectively. 
\item We propose the intermediate geometric features fusion and the language-integrated 6D pose mapping module, which integrates the geometric and linguistic information by attention mechanism and precisely maps them to the 6D pose in $SE(3)$ space. 
\item We further establish an integrated robotic system to precisely and robustly assemble blocks by natural language instructions with a 82.1\% success rate. 
\end{itemize}
\section{Related Work} \label{related}

\subsection{6D Object Pose Estimation}

6D object pose estimation has received much attention in both computer vision and robotics communities. 
A branch of approaches relies on establishing 2D-3D correspondences followed by RANSAC-based P$n$P algorithm to indirectly predict the 6D pose~\cite{rad2017bb8, peng2019pvnet, li2019cdpn, haugaard2022surfemb} while another branch directly regresses the 6D pose~\cite{xiang2018posecnn, li2018deepim, labbe2020cosypose, wang2021gdr, hu2020single, di2021so}.
In robotic manipulation tasks, \cite{zeng2017multi, tremblay2018deep, zhai2023monograspnet,zhai2023sg} leverage 6D object pose information to grasp objects. 
\cite{deng2020self} further utilize the interaction between the robot and objects to self-supervised collecting data for training 6D pose estimation network. 
Similarly, \cite{chen2022sim} proposes an iterative self-training framework for robotic bin-picking. 
\cite{wen2022catgrasp} learns task-relevant grasping for industrial objects based on category-level pose estimation~\cite{di2022gpv,zhang2022rbp,zhang2022ssp,zaccaria2023self,su2023opa}. 
On the other hand, \cite{litvak2019learning} simplifies 6D pose into 3D pose, \ie, the 2D object center and the orientation angle, and implements robotic assembly in a plane. 
\cite{stevvsic2020learning} enables robots to reason the 6D assembly pose by the surrounding geometry and generalize to the objects that contain the shape template. 
\cite{morgan_vision-driven_2021} further exploits 6D pose tracking to achieve assembly tasks with tight tolerances, including box packing, cup stacking and marker insertion. 
However, the 6D pose prediction above only relies on geometric information; thus estimating 6D poses with language instructions is beyond their reach. 

\subsection{Robotic Manipulation with Language Instructions}

The blossom of CV and NLP enables robots to comprehend natural language instructions. 
Specifically, visual grounding tasks \cite{yu2016modeling, yu2018mattnet, kamath2021mdetr} predict the 2D bounding boxes of the target objects referring to language descriptions. 
CLEVR~\cite{johnson2017clevr} contains a branch of cubes, cylinders and spheres with various attributes and provides visual reasoning, including counting, comparison and logical operations. 
Instead of asking questions about the scene as in CLEVR~\cite{johnson2017clevr}, CLEVR-Ref+~\cite{liu2019clevr} produces the bounding boxes and segmentation masks based on the given referring expressions. 
As the blossom of transformer language models \cite{devlin2018bert,liu2019roberta}, the performance of visual grounding reaches a remarkable level \cite{kamath2021mdetr}.

Some works employ visual grounding methodology to locate objects for robotic grasping. 
\cite{hatori2018interactively, ahn2018interactive, shridhar2018interactive} propose to grasp the real-world objects by natural language instructions to accomplish human-robot collaboration. 
\cite{chen2021joint} exploits LSTM to directly detect grasps from textual command rather than localizing the object, which avoids the error caused by inaccurate object localization. 
To overcome the uncertainties of human language, \cite{zhang2021invigorate} builds a partially observable Markov decision process to track the history of observations, which accomplishes interactively robotic grasping in clutter. 
However, the object is merely localized by 2D bounding boxes or 2D grasp candidates. 
2D location may be sufficient for grasping, but high-precision robotic assembly exceeds its capability. 

Recently, \cite{ahn2022can} further combines large language models with reinforcement learning to enable robots to comprehend long-horizon, abstract language instructions. 
Considering the robot's capability to follow the language instructions given continuously, \cite{ahn2022visually} enhances visual grounding tasks to history-dependent ones. 
Additionally, \cite{cheang2022learning} leverages 6D object poses to grasp category-level objects, which is instructed by natural language. 
However, only the 2D localization information is extracted within the text and the mapping between natural language and 6D object pose is not established. 

\section{Method} \label{method}

\begin{figure*}
	\centering
	\vspace{2mm}
	\includegraphics[width = 0.99\linewidth]{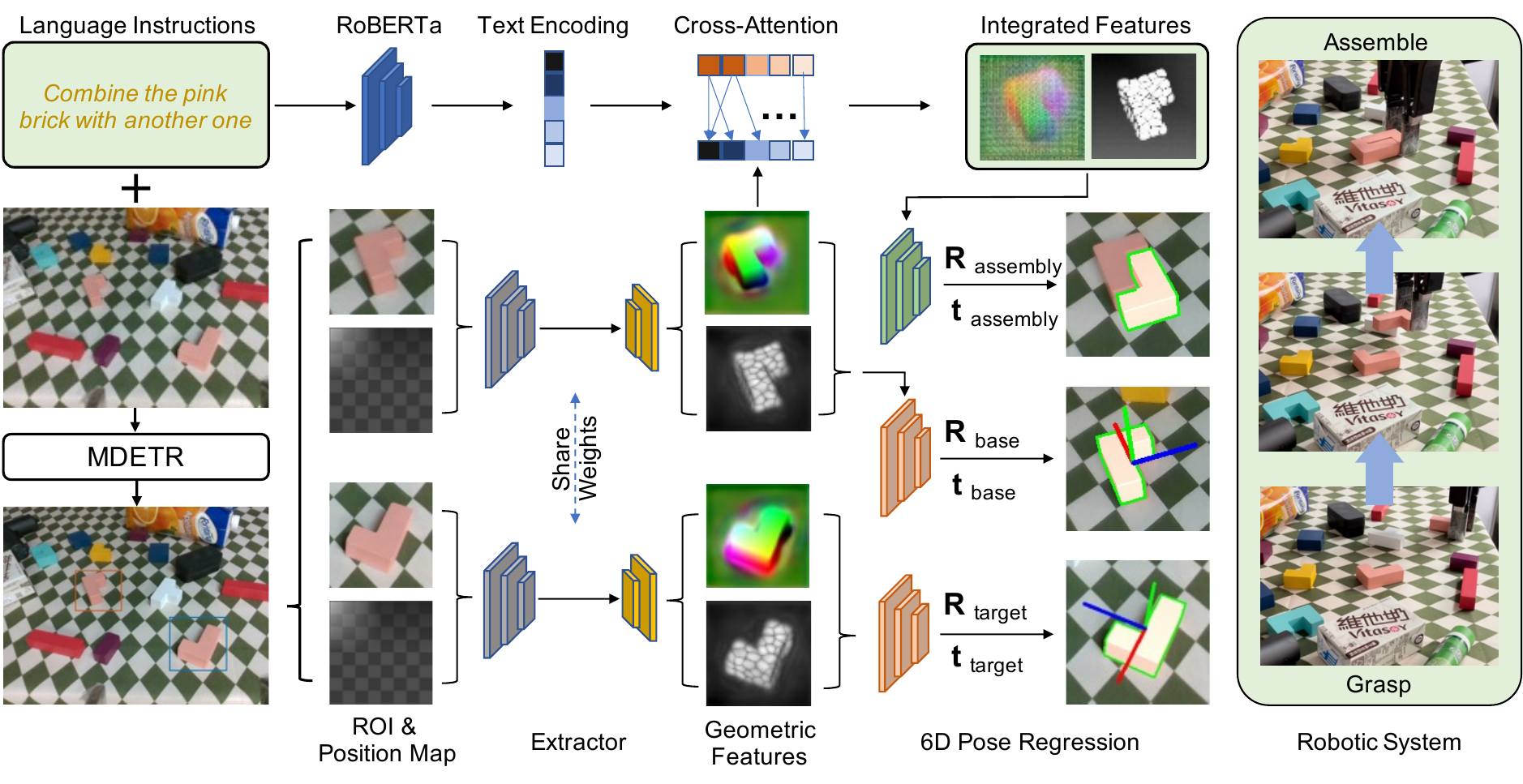}\vspace{-4mm}
	\caption{Overview of our system. Given an RGB image and a description text, we first locate the base object and the target object by a 2D visual grounding network MDETR~\cite{kamath2021mdetr}. The object ROIs are then fed into our proposed \textit{Language-Instructed 6D Pose Regression Network} (LanPose) to jointly predict the object poses and the assembly pose. Specifically, our network predicts several intermediate geometric feature maps and then directly regresses the 6D base object pose. Simultaneously, the text encoding is produced by RoBRETa,  integrated along with geometric features by a cross-attention module and mapped to the 6D assembly pose. Subsequently, an integrated robotic is established to grasp, manipulate and assemble the target block precisely to the base block. }
	\label{fig:net}
\end{figure*}

\subsection{Problem Statement}

Given an RGB image $\mathcal{I}$ and a natural language instruction $\mathcal{W}$, we aim to estimate the 6D poses of the base object $\mathcal{P}_{base}$, the target object $\mathcal{P}_{target}$ and the assembly position $\mathcal{P}_{assembly}$ and then conduct robotic assembly. 

Our framework is presented in Fig.~\ref{fig:net}. 
The RGB image $\mathcal{I}$ and the language instruction $\mathcal{W}$ are fed to a 2D visual grounding network to acquire the 2D bounding boxes of the base object $\mathcal{B}_{base}$ and the target object $\mathcal{B}_{target}$. 
\begin{equation}
    (\mathcal{B}_{base}, ~\mathcal{B}_{target}) = \mathcal{F}_1(\mathcal{I}, ~\mathcal{W})
\label{eq:F1}
\end{equation}
Then the ROI of the base object along with the language instruction $\mathcal{W}$ is fed to the \textit{Language-Instructed 6D Pose Regression Network} to jointly estimate the 6D poses of the base object $\mathcal{P}_{base}$ and the assembly position $\mathcal{P}_{assembly}$. 
\begin{equation}
    (\mathcal{P}_{base}, ~\mathcal{P}_{assembly}) = \mathcal{F}_2(\mathcal{I}, ~\mathcal{B}_{base}, ~\mathcal{W})
\label{eq:F2}
\end{equation}
Meanwhile, the target object pose $\mathcal{P}_{target}$ is predicted through the direct pose regression branch of the network. 
\begin{equation}
    \mathcal{P}_{target} = \mathcal{F}_2(\mathcal{I}, ~\mathcal{B}_{target})
\label{eq:F2_2}
\end{equation}

Subsequently, the robotic system grasps the target object with the pose $\mathcal{P}_{target}$ and assemble it to the base object with the pose $\mathcal{P}_{assembly}$. 

\subsection{Data Generation}

We generate synthetic training and validation data based on BlenderProc~\cite{denninger2020blenderproc}, a physically-based rendering methodology. 
To enhance the generalization ability, the objects with arbitrary initial poses are dropped to planes with different gradients, heights and material. 
The 2D bounding boxes and 6D poses of objects are recorded by the physics engine Pybullet and the photorealistic images are rendered via Blender, with the camera views sampled within a certain range. 

Subsequently, we follow \cite{liu2019clevr} to generate referring expressions for the rendered images, which correspond the object attributes and positions to the language description. 
Several templates with different grammatical structures (\eg, $\langle verb \rangle ~\langle obj_1 \rangle ~\langle prep \rangle ~\langle obj_2 \rangle$, $Grab ~\langle obj_1 \rangle$ $and~ \langle verb \rangle ~it$ $\langle prep \rangle ~\langle obj_2 \rangle$) and the attributes of the objects including shape and color are pre-defined. 
Additionally, we design 7 actions for robots to comprehend, each with several pre-defined phrases (\eg, $insert ~to$, $stack ~on$, $assemble ~behind$), respectively. 
See Sec.~\ref{exp-setup} for more details. 
The corresponding 6D pose label of the assembly position is calculated by multiply of the homogeneous transformation matrix of the base object and the relative transformation between base object and assembly position, which is obtained by Blender. 

In general, the ground-truth 6D poses of the observed objects in the image are required by the physics engine and the ground-truth 6D assembly poses are jointly generated with the referring expressions, which differ according to the desired actions. 
Aided by the photorealistic images along with augmented text, our networks are trained solely with synthetic data. 
With the capability of transferring directly to real scenes without further training, our methodology is highly efficient and extensible, which avoids labor-consuming real-world annotations. 

\subsection{2D Visual Grounding Network}

In order to take advantage of the accuracy and runtime progress brought by the rapidly growing 2D object detection community without altering or re-training the pose network, most 6D object pose estimation methods \cite{xiang2018posecnn, li2019cdpn, wang2021gdr} disentangle 2D object detection and 6D object pose estimation process. 
In this case, we adopt MDETR~\cite{kamath2021mdetr}, an end-to-end modulated detector which detects objects in an image instructed by a text query, to localize the target object and the base object. 
Pre-trained on 1.3M text-image pairs, MDETR shows big gains on phrase grounding, referring expression comprehension and segmentation tasks. 
We subsequently fine-tune the network on our synthetic custom dataset, using Adam optimizer with a learning rate of 5e-5. 

To sum up, the RGB image captured by the camera is fed to MDETR along with the text description to produce the 2D bounding boxes of the target object and the base object. 

\subsection{Language-Instructed 6D Pose Regression Network} \label{pose_network}

After acquiring the 2D localization of the base object and the target object, our network is required to integrate both image and text information to predict 6D poses. 
For the target object, only the 6D pose itself needs to be estimated for robotic grasping. 
However, for the base object, we would like to predict the corresponding assembly position as well. 
We design \textit{Language-Instructed 6D Pose Regression Network}, namely LanPose, to simultaneously predict the 6D poses of both base object and assembly position, as presented in Fig.~\ref{fig:net}. 

\textbf{Direct 6D pose regression branch.~~}
Given the ROI of the base object, we directly regress its 6D pose. 
Specifically, several intermediate geometric feature maps that indicate the 2D-3D correspondences are predicted to supervise the network. 
Then a CNN module, which supersedes the traditional RANSAC-based P$n$P algorithm, is leveraged to directly regress the 6D pose $\mathcal{P}_{base}$. 
Noteworthy, the geometric feature maps also imply the surrounding context of the target assembly position, which is crucial to the 6D pose prediction of the assembly position. 

\textbf{Language-integrated 6D pose mapping branch.~~}
On the other hand, the text input is fed into a pre-trained transformer language model RoBERTa~\cite{liu2019roberta} to yield a sequence of hidden vectors, which is the same size as the text input. 
Then the geometric feature maps, which contain the pose and shape context of the observed object, and the text encoding, which indicates the relative relationship of target assembly position towards the observed object, are fused to produce an integrated image and text features. 
We exploit cross-attention mechanism~\cite{vaswani2017attention} to integrate image and language features, which computes attention scores between the query, which in this case is the geometric feature maps, and the key and value, which is the text encoding. 
These attention scores are utilized to weight the values, which are then aggregated to produce a joint representation of the integrated text-image feature.
Additionally, a CNN module is exploited to subsequently predict the 6D assembly position pose w.r.t. the camera $\mathcal{P}_{assembly}$.
Specifically, the CNN module consists of three convolutional layers, each followed by Group Normalization and GELU activation. 
Two Fully Connected (FC) layers are then applied to the global feature. 
Finally, two parallel FC layers output the 3D rotation $\rot$ and 3D translation $\trans$, respectively. 

\textbf{Losses.~~}
Inspired by \cite{li2019cdpn}, we disentangle 6D assembly position pose loss as
\begin{equation}
    \loss_{Assembly} = \loss_\rot + \loss_{center} + \loss_z.
\label{eq:loss_pose}
\end{equation}
We employ point matching loss for rotation and parametrize rotation as the first two columns of the rotation matrix following \cite{wang2021gdr}. 
The translation is further decoupled to the 2D center location and the distance w.r.t. the camera. 
Therefore, 
\begin{equation}
    \begin{cases}
        \loss_\rot &= \underset{\mathbf{x} \in \mathcal{M}}{\avg} \| \hat{\rot} \mathbf{x} - \bar{\rot} \mathbf{x} \|_1 \\
        \loss_{\text{center}} &= \| (\hat{\Delta}_x - \bar{\Delta}_x,  \hat{\Delta}_y - \bar{\Delta}_y) \|_1 \\
        \loss_{z} &= \| \hat{\Delta}_z - \bar{\Delta}_z \|_1
    \end{cases}, 
\label{eq:loss_pose_detail}
\end{equation}
where $\mathcal{M}$ represents object model vertices, $\Delta$ indicates the scale-invariant translation~\cite{li2019cdpn} and $\hat{\bullet}$ and $\bar{\bullet}$ severally denote estimation and ground-truth. 
For symmetric objects, we rotate $\bar{\rot}$ along the known symmetry axes to the best match of $\hat{\rot}$. 
The overall loss of LanPose can be summarized as
\begin{equation}
    \loss_{LanPose} = \loss_{Pose} + \loss_{Geom} + \loss_{Assembly}, 
\label{eq:loss_total}
\end{equation}
where $\loss_{Pose}$ and $\loss_{Geom}$ are defined in GDR-Net~\cite{wang2021gdr}. 

\textbf{Training strategy.~~}
To enhance the precision and robustness of the network, the \textit{direct 6D pose regression branch} is required to solely reflect the geometric context of the observed object without the effect of language instructions. 
Thereby, we train \textit{direct 6D pose regression branch} individually in stage 1. 
Then the backbone, geometric heads and pose regression module are frozen in stage 2 to provide surrounding features of the assembly position for the \textit{language-integrated 6D pose mapping branch}. 
Experimentally, we train the network for 120 epochs in stage 1 and 100 in stage 2, using Ranger optimizer with a batch size of 128 and a base learning rate of 1e-4. 


\section{Experiments} \label{exp}

\begin{figure*}[ht]
	\centering
    \vspace{2mm}
	\includegraphics[width = 0.99\linewidth]{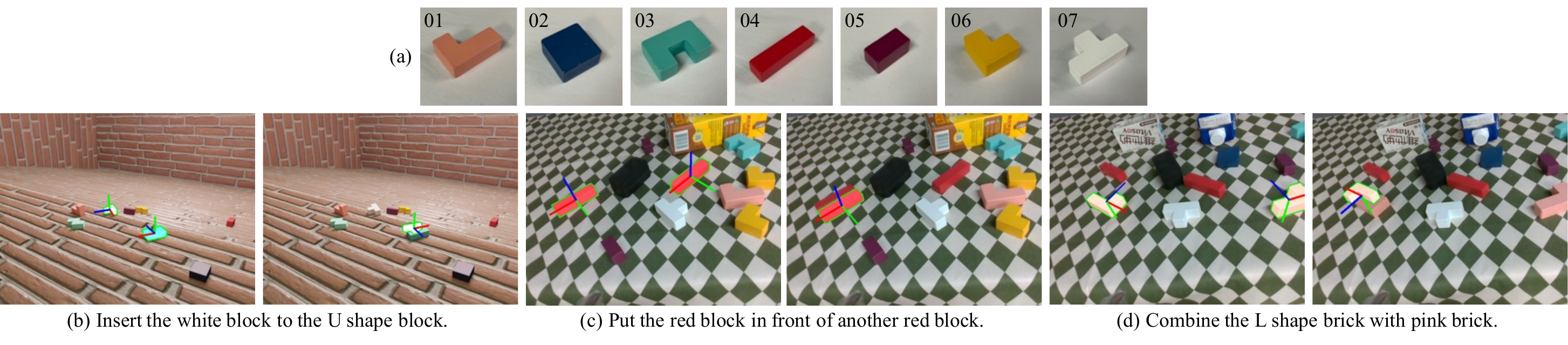}
	\vspace{-2mm}
    \caption{Visualization. (a) The 7 types of blocks we use. (b, c, d) Three groups of language-instructed 6D pose regression results by rendering the 3D models and overlaying the contours and coordinate axes on the image. Left: The 6D poses of the target object and the base object. Right: The 6D pose of assembly position. Noteworthy, the blocks are placed on planes with various gradients and heights. }
	\label{fig:visualization}
\end{figure*}

\subsection{Experimental Setup} \label{exp-setup}

We generate 39000 synthetic images for training and 1000 for validation (Fig.~\ref{fig:visualization}(b)).
7 types of blocks are employed for language-instructed 2D localization, 6D pose estimation and real-world robotic assembly experiments. 
Meanwhile, 7 assembly actions are designed, including $insert ~to$, $combine ~with$, $stack ~on$ and $assemble$ to the $front$, $right$, $left$ or $back$ of the object, each with various descriptions (\eg, $combine ~with$ \& $assemble ~together ~with$, $stack ~on$ \& $put ~upon$). 
The corresponding ground-truth assembly poses are designed to closely fit the base object to conduct high-precision robotic assembly. 

\subsection{2D Visual Grounding}

Three backbones are employed for MDETR, as shown in Table.~\ref{tab:MDETR_result}. 
As can be seen, 0.794, 0.734 and 0.815 in Average Precision (AP) is derived with ResNet-101, EfficientNet-B3 and EfficientNet-B5 backbones, respectively. 
Practically, we exploit ResNet-101 backbone in our robotic assembly experiments as a trade-off between accuracy and runtime. 
Two bounding boxes with the highest confidence are selected for language-instructed 6D pose inference along with the corresponding object categories. 

\begin{table}[b]
    \centering
    \caption{Results for 2D Visual Grounding}
    \label{tab:MDETR_result}
    \setlength{\tabcolsep}{4mm}
    \begin{tabular}{cccccc}
    \toprule
         MDETR Backbone & AP & AP50 & AP75 \\
    \midrule
         ResNet-101       & 0.794 & 0.990 & 0.930 \\
         EfficientNet-B3  & 0.734 & 0.990 & 0.986 \\
         EfficientNet-B5  & 0.815 & 0.990 & 0.973 \\
    \bottomrule
    \end{tabular}

\end{table}
 
\subsection{Language-Instructed 6D Pose Inference}

\begin{table}[t]
    \centering
    \caption{Results for 6D pose estimation on validation set}
    \label{tab:6D_pose}
    \setlength{\tabcolsep}{3.5mm}
    \begin{tabular}{cccccc}
    \toprule
        \multirow{2}{*}{Block} & \multicolumn{3}{c}{ADD(-S)} & \multirow{2}{*}{2$^\circ$ 2cm} & \multirow{2}{*}{5$^\circ$ 5cm} \\
                                 & 0.02d & 0.05d & 0.1d       & & \\
    \midrule
        \multicolumn{6}{c}{6D Object Pose} \\
    \midrule
         01 & 74.32 & 95.33 & 98.19 & 85.08 & 95.18\\
         02 & 56.76 & 91.49 & 98.50 & 69.97 & 91.24\\
         03 & 74.72 & 95.30 & 98.65 & 85.24 & 95.90\\
         04 & 75.40 & 94.03 & 97.89 & 80.42 & 92.42\\
         05 & 43.73 & 84.10 & 96.34 & 72.52 & 92.73\\
         06 & 58.59 & 93.57 & 98.54 & 85.98 & 96.73\\
         07 & 64.10 & 93.31 & 98.49 & 81.80 & 95.42\\
         \textbf{Mean} & \textbf{63.95} & \textbf{92.45} & \textbf{98.09} & \textbf{80.14} & \textbf{94.23} \\
    \midrule
        \multicolumn{6}{c}{6D Assembly Pose} \\
    \midrule
         01 & 57.24 & 91.85 & 98.09 & 45.67 & 92.56\\
         02 & 27.40 & 67.40 & 88.00 & 22.00 & 74.60\\
         03 & 46.64 & 93.08 & 99.40 & 55.87 & 93.18\\
         04 & 32.90 & 75.63 & 91.88 & 24.17 & 78.84\\
         05 & 20.90 & 54.37 & 83.02 & 24.42 & 84.12\\
         06 & 34.27 & 80.30 & 96.18 & 40.70 & 93.17\\
         07 & 37.84 & 84.58 & 98.30 & 56.66 & 95.30\\
         \textbf{Mean} & \textbf{36.74} & \textbf{78.17} & \textbf{93.55} & \textbf{38.50} & \textbf{87.39} \\
    \bottomrule
    \end{tabular}
\end{table}

\textbf{Evaluation metrics.~~}
We exploit ADD(-S) and \textit{n}$^\circ$ \textit{n} cm to evaluate the 6D object pose and 6D assembly pose inference performance. 
The ADD metric~\cite{hinterstoisser2012model} calculates the mean distance between the model points transformed by the predicted pose and ground-truth pose. 
For symmetric objects, the ADD-S metric~\cite{hodavn2016evaluation} measures the mean distance from the closest points in two transformed sets. 
The \textit{n}$^\circ$ \textit{n} cm measures whether the rotation error is less than \textit{n}$^\circ$ and the translation error is under \textit{n} cm. 
The minimum error among all possible ground-truth poses is taken for symmetric objects. 

\textbf{Analysis.~~}
Table.~\ref{tab:6D_pose} demonstrates the performance of our proposed LanPose. 
For the \textit{direct 6D pose regression branch}, we achieve 98.09 in ADD(-S)-0.1d and 94.23 in 5$^\circ$ 5 cm. 
With regards to more rigorous metrics ADD(-S)-0.05d and 2$^\circ$ 2 cm, 92.45 and 80.14 are derived, respectively. 
On the other hand, 93.55 in ADD(-S)-0.1d and 87.39 in 5$^\circ$ 5 cm are attained for the assembly position pose in \textit{Language-integrated 6D pose mapping branch}. 
The lack of explicit geometric features and the complexity of mapping the implicit geometric and linguistic encoding to the 6D pose space prompt the difficulty in obtaining a competitive performance with the observed object, yet it is sufficient for the robotic assembly task.
Additionally, the qualitative visualization results of real-world data are shown in Fig.~\ref{fig:visualization}(c, d), which demonstrate the high precision of our proposed network. 

\subsection{Inference Time Analysis}
With an AMD Ryzen 7 5800 CPU and an NVIDIA RTX 3090 GPU, our approach takes 115 ms to integrate image and text input and produce 2D bounding boxes, and 36 ms for jointly predicting the 6D poses of the observed object and the corresponding assembly position. 

\subsection{Robot Assembly Experiments}

\begin{figure*}[ht]
	\centering
    \vspace{2mm}
	\includegraphics[width = 0.99\linewidth]{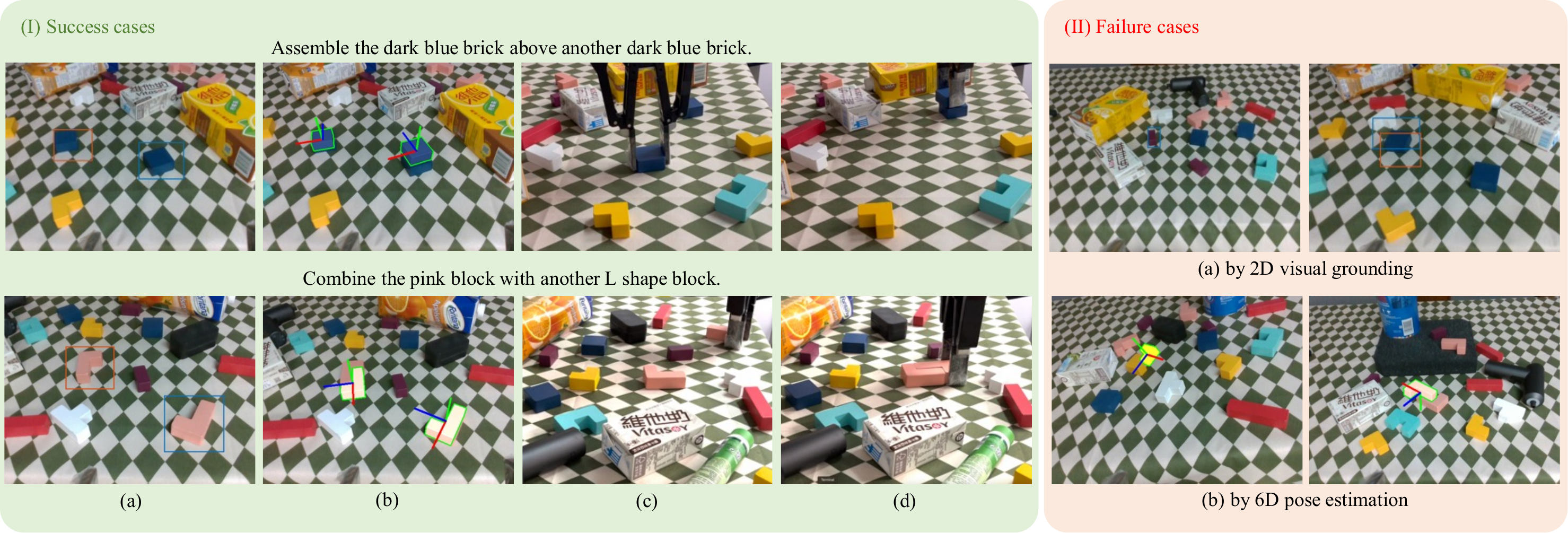}
	\vspace{-2mm}
    \caption{Visualization of robotic assembly experiments. (\uppercase\expandafter{\romannumeral1}) Success cases. (a) The 2D bounding boxes produced by visual grounding network. (b) The corresponding 6D pose estimation results of target object and assembly position. (c, d) The robotic grasping and assembly process, respectively. (\uppercase\expandafter{\romannumeral2}) Failure cases. (a) caused by 2D visual grounding. (b) caused by 6D pose estimation. }
	\label{fig:assembly_exp}
\end{figure*}
\begin{table}[t]
    \centering
    \caption{Success rate for robotic assembly experiments}
    \label{tab:success_rate}
    \setlength{\tabcolsep}{3.5 mm}
    \begin{tabular}{ccccc}
    \toprule
         Block & \makecell{2D Visual Grounding} & \makecell{Robotic Assembly} \\
    \midrule
         01 & 100.0\% & 100.0\% (20/20) \\
         02 & 90.0\% & 85.0\% (17/20) \\
         03 & 85.0\% & 80.0\% (16/20) \\
         04 & 90.0\% & 80.0\% (16/20) \\
         05 & 90.0\% & 85.0\% (17/20) \\
         06 & 75.0\% & 65.0\% (13/20) \\
         07 & 85.0\% & 80.0\% (16/20) \\
    \midrule
         Mean & 87.9\% & 82.1\% (115/140) \\
    \bottomrule
    \end{tabular}
\end{table}

We adopt KUKA LBR iiwa R820 robot arm and 2-finger Robotiq 2F-140 gripper to establish the robotic assembly system, with an Intel RealSense D435i camera mounted to the gripper. 
Robot experiments are executed by Robotic Operating System (ROS), with iiwa stack~\cite{hennersperger2017towards} communicating between ROS and KUKA Sunrise software. 

\textbf{Analysis.~~}
We conduct 20 robotic assembly trials for each block. 
Specifically, the blocks are placed with arbitrary poses in the cluttered background on the plane with various gradients and heights, and the desired assembly position is randomly given by natural language. 
As illustrated in Table.~\ref{tab:success_rate}, 87.9\% detection rate is achieved, \ie, the base and target objects are correctly located by the 2D visual grounding network. 
We consider the trial with the assembly error under 5 mm a successful one. 
Thus, 115 out of 140 trials are successful, leading to a 82.1\% total success rate of perception, grasping, manipulation and assembly, as shown in Fig.~\ref{fig:assembly_exp}(\uppercase\expandafter{\romannumeral1}). 
In real-world experiments, various language expressions sharing the same purpose are employed and repeatedly fed into the system, along with the arbitrarily placed blocks. 
Our networks robustly locate the blocks and precisely predict the 6D assembly pose, which is impregnable to the initial condition, proving that the linguistic information is finely fused with geometric features and properly mapped to the 6D pose space. 

\textbf{Failure cases.~~}
The failures are provoked by two main reasons, the 2D visual grounding failure and the 6D pose estimation error, as enumerated in Fig.~\ref{fig:assembly_exp}(\uppercase\expandafter{\romannumeral2}). 
The objects in a cluttered background or stacked together with occlusion may prompt detection mistakes. 
Accordingly, the inaccurate 2D bounding boxes and pose ambiguity caused by occlusion or symmetry lead to 6D pose estimation error. 
Practically, the 6D pose estimation precision required by robotic grasping is not as strict as that demanded by robotic assembly. 
Thus the robotic implementation failure regularly occurs in the assembly process. 

\subsection{Ablation Study}

\begin{table}[t]
    \centering
    \caption{6D assembly pose results of ablation study}
    \label{tab:ablation}
    \setlength{\tabcolsep}{1.5mm}
    \begin{tabular}{cccccc}
    \toprule
          \multirow{2}{*}{Method} & \multicolumn{3}{c}{ADD(-S)} & \multirow{2}{*}{2$^\circ$ 2cm} & \multirow{2}{*}{5$^\circ$ 5cm} \\
                                 & 0.02d & 0.05d & 0.1d       & & \\
    \midrule
         \textbf{Ours} & \textbf{36.74} & \textbf{78.17} & \textbf{93.55} & \textbf{38.50} & \textbf{87.39} \\
         w/o cross-attention & 21.93 & 63.94 & 88.75 & 15.09 & 65.61 \\
         $\mathcal{P} \rightarrow \Delta \mathcal{P}$ & 11.16 & 37.16 & 48.37 & 21.94 & 77.17 \\
    \bottomrule
    \end{tabular}

\end{table}

\textbf{Ablation on cross-attention mechanism.~~}
We compare the performance of geometric feature fusion methodology, as presented in Table.~\ref{tab:ablation}. 
Instead of exploiting the cross-attention mechanism to integrate image and language features, we directly concatenate intermediate geometric feature maps and text encoding in \textit{language-integrated 6D pose mapping branch}. 
The experimental results illustrate that the cross-attention mechanism surpasses the direct concatenation of explicit geometric feature maps and plays a vital role on modeling multi-modality relationships. 

\textbf{Ablation on pose type.~~}
In the \textit{language-integrated 6D pose mapping branch}, we fuse text encoding and geometric feature maps to predict the absolute 6D assembly pose w.r.t. the camera. 
As demonstrated in Table.~\ref{tab:ablation}, we compare the results of predicting the relative 6D assembly pose w.r.t. the base object with the baseline. 
Although the language description reflects the relative position between the observed object and the desired assembly position, the quantitative results of predicting $\Delta \mathcal{P}_{assembly}$ are not competitive with directly predicting $\mathcal{P}_{assembly}$, which suggests the explicit surrounding context provided by intermediate geometric maps dominates the prediction of 6D assembly pose. 

\section{Conclusion} \label{conclusion}

In this paper, we establish an integrated robotic system to robustly assemble blocks by natural language instructions. 
To this aim, \textit{Language-Instructed 6D Pose Regression Network} (LanPose) is proposed, which integrates image and language information and maps them to the 6D pose in $SE(3)$ space. 
With the two-stage training strategy, the 6D poses of the observed object and the corresponding assembly position are jointly predicted without being coupled. 
Meanwhile, the integration of intermediate geometric feature maps and text encoding by the attention mechanism and the language-integrated 6D pose mapping module enormously enhance the precision of the network. 

Along with the reliable 2D visual grounding network, our system is capable of perceiving, grasping, manipulating and assembling blocks robustly. 
Aided by the photorealistic images generated by physically-based rendering and the augmented text description, our networks trained solely with synthetic data are able to transfer directly to real-world applications without further fine-tuning. 
Our proposed methodology demonstrates the potential to more efficient human-robot cooperation with natural language instructions. 

\bibliographystyle{IEEEtran}
\bibliography{ref}

\end{document}